\documentclass[10pt,twocolumn,letterpaper]{article}

\usepackage{iccv}
\usepackage{times}
\usepackage{epsfig}
\usepackage{graphicx}
\usepackage{amsmath}
\usepackage{amssymb}
\usepackage{dblfloatfix} 
\usepackage{caption}

\usepackage[pagebackref=true,breaklinks=true,letterpaper=true,colorlinks,bookmarks=false]{hyperref}

\DeclareMathOperator*{\argmin}{arg\,min}

\iccvfinalcopy 

\def\iccvPaperID{23} 

\ificcvfinal\fi

\begin{document}

\title{Deep Hyperspectral Prior: \\ Single-Image Denoising, Inpainting, Super-Resolution}

\author{First Author\\
Institution1\\
Institution1 address\\
{\tt\small firstauthor@i1.org}
\and
Second Author\\
Institution2\\
First line of institution2 address\\
{\tt\small secondauthor@i2.org}
}

\author{Oleksii Sidorov and Jon Hardeberg\\
The Norwegian Colour and Visual Computing Laboratory, NTNU\\
Gj\o vik, Norway\\
{\tt\small oleksiis@stud.ntnu.no, jon.hardeberg@ntnu.no}
}

\ificcvfinal\thispagestyle{empty}\fi

\twocolumn[{%
\renewcommand\twocolumn[1][]{#1}%
\maketitle
\vspace{-\baselineskip}
    \includegraphics[width=\linewidth]{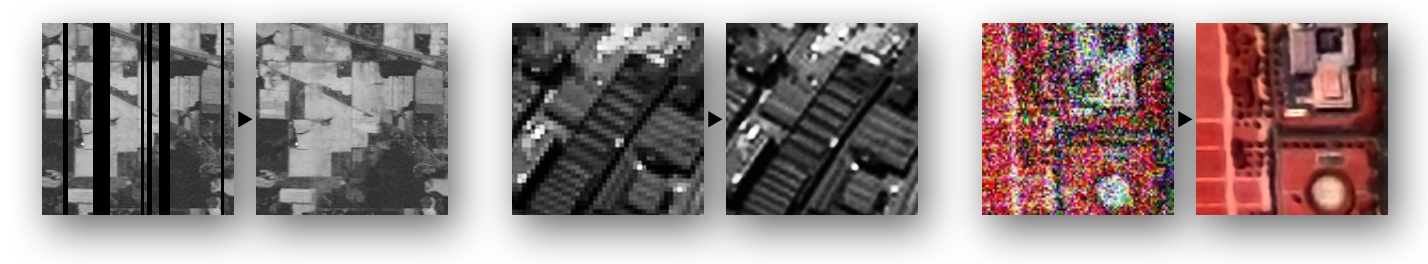}
    \vspace{\baselineskip}%
}]
\begin{abstract}
Deep learning algorithms have demonstrated state-of-the-art performance in various tasks of image restoration. This was made possible through the ability of CNNs to learn from large exemplar sets. However, the latter becomes an issue for hyperspectral image processing where datasets commonly consist of just a few images. In this work, we propose a new approach to denoising, inpainting, and super-resolution of hyperspectral image data using intrinsic properties of a CNN without any training. The performance of the given algorithm is shown to be comparable to the performance of trained networks, while its application is not restricted by the availability of training data. This work is an extension of original ``deep prior" algorithm to hyperspectral imaging domain and 3D-convolutional networks.
\end{abstract}

\section{Introduction}
Deep Convolutional Neural Networks (CNNs) are occupying more and more leading positions in benchmarks of various image processing tasks. Commonly, it is related to the excellent representative ability of hierarchical convolutional layers which allows CNNs to learn a large amount of visual data without any hand-crafted assumptions. Ulyanov \etal \cite{ulyanov2018deep} were the first to show that not only a learning ability but also the inner structure of a CNN itself can be beneficial for processing of image data.

For example, the inverse task of image restoration, such as inpainting, noise removal, or super-resolution, can be formulated as an energy minimization problem as follows:
\begin{equation}
    x^*=\min_xE(x,x_0)+R(x)\quad,
\end{equation}
where $E(x,x_0)$ is a task related metric, $x$ and $x_0$ are original and corrupted images, and $R(x)$ is a regularization term (image prior) which can be chosen manually or can be learned from data (as it happens in the vast majority of CNN-based methods). However, the theory of Ulyanov \etal states that image prior can be found in the space of the network's parameters directly, through the optimization process, which allows removal of regularization term, and searching a solution as:
\begin{equation}
    x^*=f_{\theta^*}(z), \quad where \,\,\, \theta^*=\argmin_xE(f_\theta(z),x_0)
\end{equation}
Here, $f_θ$ is a CNN with parameters $θ$, and $z$ is a fixed input (noise). Thereby, an original image can be restored via optimization of the network's weights using only a corrupted image.

This approach has a particularly high significance in the domain of hyperspectral imaging (HSI). Currently, HSI is a powerful tool which is widely used in remote sensing, agriculture, cultural heritage, food industry, pharmaceutics, \emph{etc}. The complexity of hyperspectral equipment and process of data acquisition make corruption of image data even more likely than it is for RGB imaging. Thus, it generates an increased demand for algorithms of hyperspectral image restoration. But, at the same time, accurate learning-based methods can hardly be used due to the lack of data. The complexity of data acquisition does not allow gathering of large custom datasets for a particular task, and even openly available ones are very limited and rarely exceed one hundred images, sometimes consisting of just one image \cite{PURR1947}\cite{fauvel2012advances}\cite{wang2014hyperspectral}.  

Our work aims to solve this problem and, altogether, our contributions can be formulated as follows:
\begin{itemize}
    \item We propose an efficient algorithm for hyperspectral image restoration based on the theory of Ulyanov \etal
    \item We design a new 3D-convolutional implementation of the algorithm and prove the fundamental property of 3D convolutions to contain low-level image information which can be used as a prior.
    \item We demonstrate a potential application of the algorithm for HSI and evaluate its performance in comparison with other methods. 
    \item Eventually, we make the source code publicly accessible and is ready to use out-of-the-box.
\end{itemize}
	
\begin{figure*}[b!]
\begin{center}
\includegraphics[width=\linewidth]{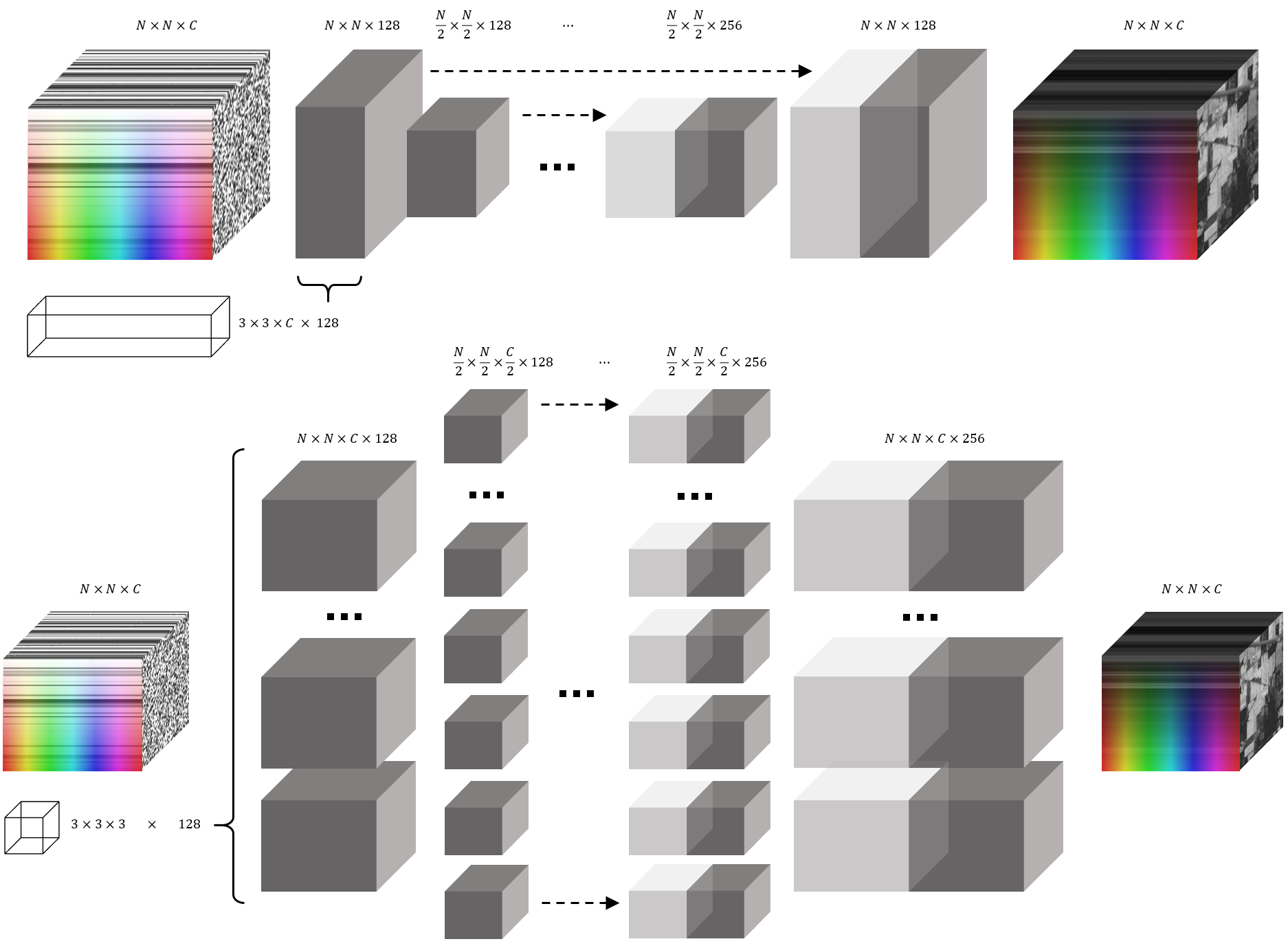}
\end{center}
   \caption{2D (top) and 3D (bottom) convolutional architectures used in our experiments. Size of the filters of the first convolutional layer is illustrated under the input image. Pooling and activation layers are omitted for simplicity sake.}
\label{fig:2}
\end{figure*}

\section{Related Works}
In this section, we briefly introduce recent advances in hyperspectral denoising, inpainting, and super-resolution. 

One way to perform HSI denoising is to apply 2D algorithms to each band separately. Such an approach can utilize bilateral \cite{Tomasi:1998:BFG:938978.939190} or NL-means filtering \cite{buades2005non}, total variation \cite{rudin1992nonlinear}, block-matching 3D-filtering \cite{dabov2007video}, or novel CNN-based techniques, \eg DnCNN \cite{Zhang:2017:BGD:3101312.3101324}. However, not taking spectral data into consideration may cause distortions and artifacts in the spectral domain. This, has given rise to a family of algorithms based on spatial-spectral features, such as spatiospectral derivative-domain wavelet shrinkage \cite{1580725}, low-rank tensor approximation \cite{4455564}, low-rank matrix recovery \cite{6648433}, and most recently FastHyDe algorithm \cite{8289385}, which utilizes sparse representation of an image linked to its low-rank and self-similarity characteristics. A deep learning paradigm has been used in the 3D modification of DnCNN, and more advanced HSI-oriented network HSID-CNN \cite{8454887}.

The inpainting of grayscale and RGB images conventionally rely on patch-similarity and variational algorithms to propagate information from intact regions to holes \cite{Barnes:2009:PAR}\cite{bertalmio2000image}\cite{efros2001image} and may be used for HSI data in a band-wise manner. Novel inpainting approaches benefit from realistic reconstruction ability of GANs, which allows filling even large holes with remarkable accuracy \cite{IizukaSIGGRAPH2017}\cite{liu2018image}\cite{8578675}. However, they rely on large training datasets. There are also a number of HSI-specific inpainting methods \cite{bousefsaf2018image}\cite{article1}\cite{degraux2015generalized}. Similar to our approach, Addesso \etal \cite{8297090} address HSI inpainting as an optimization task with (hand-crafted) collaborative total variation regularizer, while Yao \etal \cite{4_2017_zhuang_HyInpaint} designed a regularizer based on the Criminisi's inpainting method. Recently, the FastHyIn algorithm \cite{8289385} (an extension of FastHyDe) demonstrated state-of-the-art HSI inpainting accuracy, with the only remark that similar to \cite{6484934} it utilizes information from intact bands, thus cannot be used in cases of all-bands corruption.

The majority of hyperspectral super-resolution (SR) algorithms perform a fusion of input hyperspectral image with a high-resolution multispectral image which is easier to obtain \cite{8099894}\cite{qu2018unsupervised}. Single-image SR is a more sophisticated task. The attempts to solve it include spectral mixture analysis \cite{7517322}, low-rank tensor approximation \cite{wang2017hyperspectral}, Local-Global Combined Network \cite{7937881}, MRF-based energy minimization \cite{8310636}, transfer learning \cite{7855724}, the recent method of 3D Full Convolutional Neural Network \cite{mei2017hyperspectral}, and others.

\section{Methodology}
The main idea of the method is captured by Equations (1) and (2). The fully-convolutional encoder-decoder $f_\theta$ is designed to translate a fixed input $z$ filled with noise to the original image $x$, conditioned on corrupted image $x_0$. We use the paradigm of the ``deep prior" method \cite{ulyanov2018deep} which says that optimal weights $\theta$ of a network $f_\theta$ can be found from the intrinsic prior contained in a network structure instead of learning them from the data. Particularly, $\theta$ is approximated by the minimizer $\theta^*$:
\begin{equation}
    \theta^*=\argmin_xE(f_\theta(z),x_0)
\end{equation}
which can be obtained using an optimizer such as gradient descent from randomly initiated parameters. It is also possible to optimize over input $z$ (not covered in this work). 

The energy function $E(x,x_0)$ may be chosen accordingly to the application task. In the case of a basic reconstruction problem, it may be formulated as $L_2$-distance:
\begin{equation}
    E(x,x_0)=||x-x_0||^2
\end{equation}
It was shown \cite{ulyanov2018deep} that optimization converges faster in cases of natural-looking images rather than random noise, \emph{i.e.} the process demonstrates high impedance to noise and low impedance to signal. The latter can be used to interrupt the reconstruction before the noise will be recovered, which will lead to \emph{blind image denoising}.

Also, $E(x,x_0)$ term can be modified to fill the missing regions in a \emph{inpainting} problem with mask $m\in\{0,1\}$:
\begin{equation}
    E(x,x_0)=||x-x_0\circ m||^2
\end{equation}
where $\circ$ is a Hadamard product. Otherwise, downsampling operator $d(x,\alpha):x^{\alpha N\times\alpha N\times C}\rightarrow x^{N\times N\times C}$ with factor $\alpha$ can be used in $E(x,x_0)$ to address \emph{super-resolution} task as a prediction of high-res image $x$ which, when downsampled, is the same as low-res image $x_0$:
\begin{equation}
    E(x,x_0)=||d(x)-x_0||^2
\end{equation}

\subsection{Implementation details}
It was found, that different fully-convolutional encoder-decoder architectures (sometimes with skip-connections) are suitable for the implementation of the given method. For the exact description in details please see the source code\footnote{\url{https://github.com/acecreamu/deep-hs-prior} }. Although parameters differ for each sub-task, the general framework is common and is illustrated in Fig. \ref{fig:2}.

We experiment with two versions of the networks -- 2D and 3D. While 2D convolutions are still able to process multi-channel input, they cause shrinkage of spectral information already at the first convolutional layer and recover it back at the last one. In this case, filters of these layers would have an ``elongated" shape with a depth equal to the depth of the hyperspectral image. A 3D convolution allows the use of smaller filters (\eg $3\times3\times3$) along the whole network because its output is a 3D volume. This ability to preserve a 3D shape of the input is considered to be beneficial for the processing of hyperspectral data. It is worth mentioning, that unlike a conventional ``hourglass" architecture, where data is consecutively downsampled/upsampled along two spatial dimensions, processing of 3D volumes allows doing the same with third dimension as well (see Fig. \ref{fig:2}). 

\begin{figure*}[t]
\begin{center}
\includegraphics[width=\linewidth]{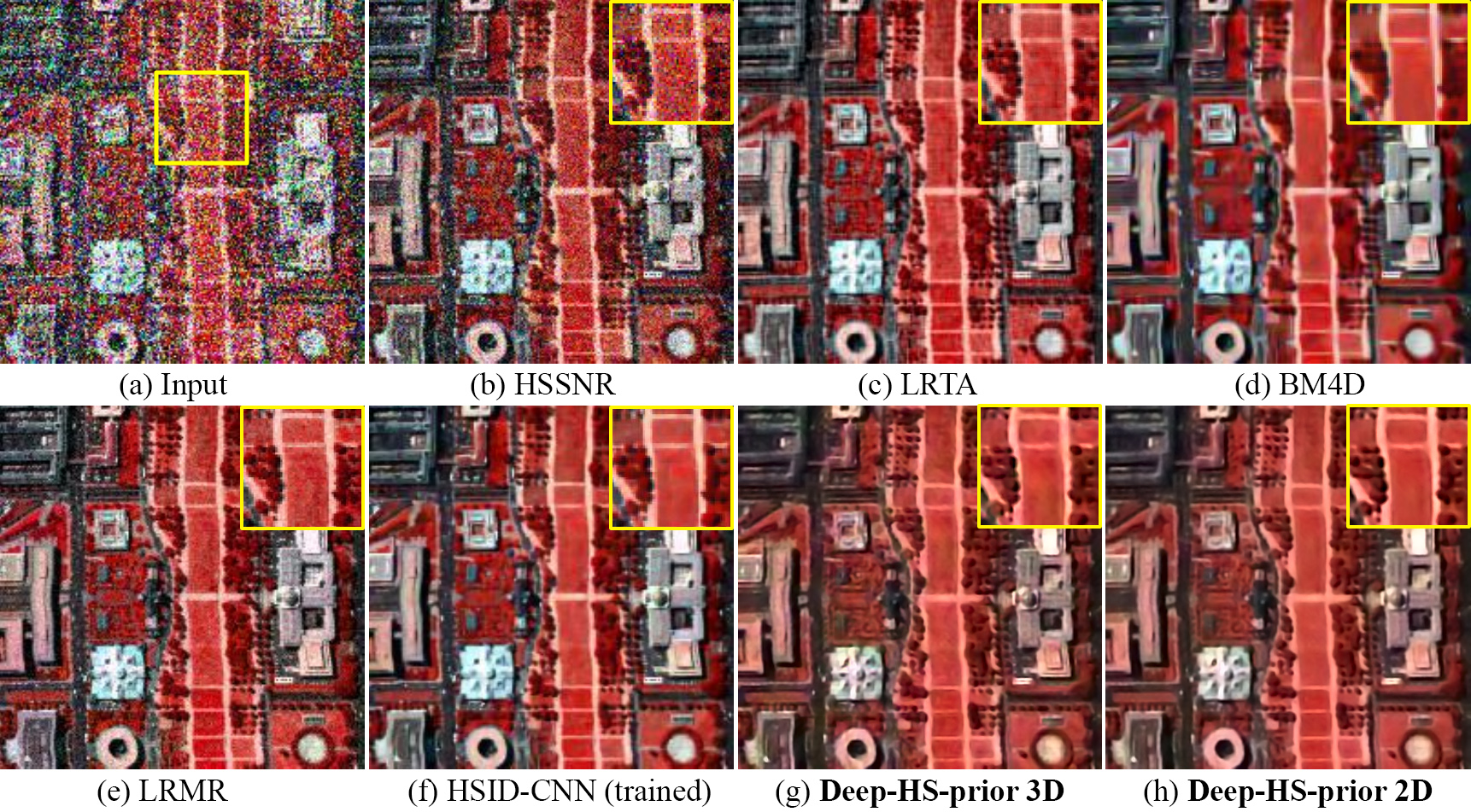}
\end{center}
   \caption{HSI denoising results. HYDICE DC Mall image; false-color with bands (57, 27, 17).}
\label{fig:3}
\end{figure*}

The input of the network is uniform noise in range 0-0.1 of a shape equal to the shape of a processed hyperspectral image. Optionally, it is additionally perturbed at each iteration with Gaussian noise of specified $\sigma$. The activation function used is LeakyReLU \cite{xu2015empirical}. Downsampling is performed using a stride of convolutions, while upsampling is either ``nearest" or bilinear (trilinear for a 3D case). Other methods can also be used but these prevailed in our experiments. ADAM algorithm was used for optimization.

\section{Experimental setup}
\paragraph{Denoising.} 
We evaluate the ability of an algorithm to remove noise using HYDICE DC Mall data \cite{wang2014hyperspectral} with synthetically added Gaussian noise of $\sigma=100$. The image consists of 191 channels and was cropped to $200\times200$ pixels size. Results (Fig. \ref{fig:3}, Table \ref{tab:1}) are compared to HSSNR \cite{1580725}, LRTA \cite{4455564}, BM4D \cite{6253256}, LRMR \cite{6648433}, and HSID-CNN \cite{8454887} methods.

\begin{figure*}[]
\begin{center}
\includegraphics[width=\linewidth]{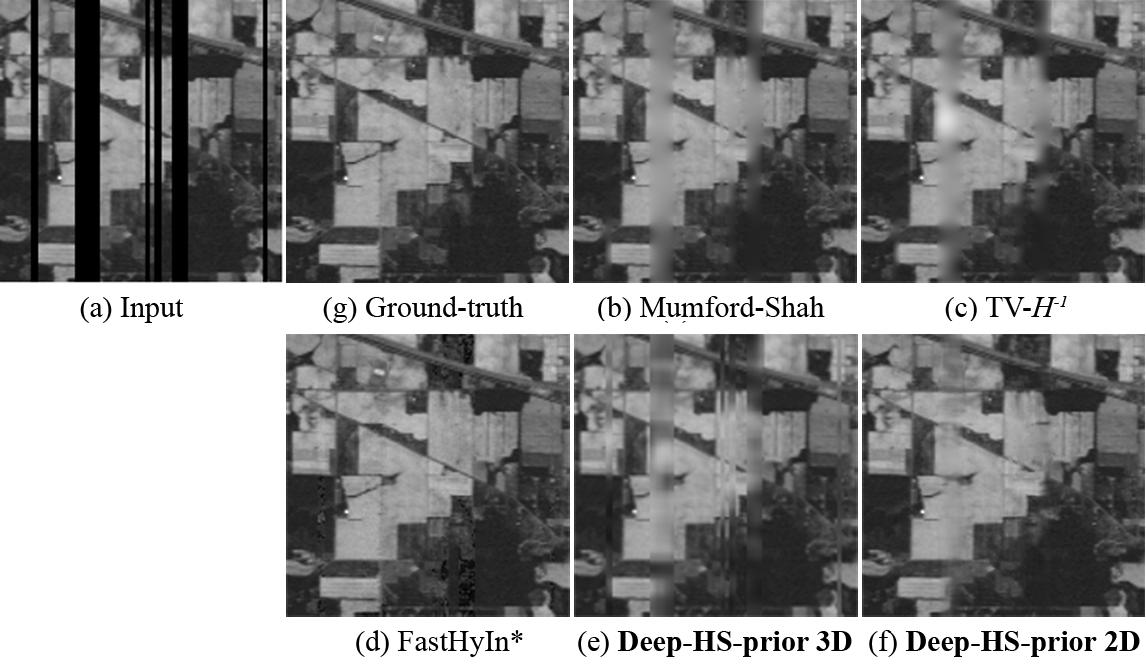}
\end{center}
   \caption{HSI inpainting results. AVIRIS Indian Pines dataset; band 150. The mask of corrupted stripes (a) was applied to all bands, except case (d) where only bands 25:175 (75\%) were affected.}
\label{fig:4}
\end{figure*}

\paragraph{Inpainting.}
The Indian Pines dataset \cite{PURR1947} ($145\times145\times200$) from AVIRIS  sensor was used to test the proposed inpainting method. The mask of corrupted strips was applied to all bands. Results (Fig. \ref{fig:4}, Table \ref{tab:1}) are compared to Mumford-Shah \cite{esedoglu2002digital} and fourth-order total variation (TV-$H^{-1}$) \cite{bertozzi2011unconditionally} 2D methods, as well as the state-of-the-art HSI inpainting method FastHyIn \cite{8289385}. 

\begin{figure*}[b]
\begin{center}
\includegraphics[width=\linewidth]{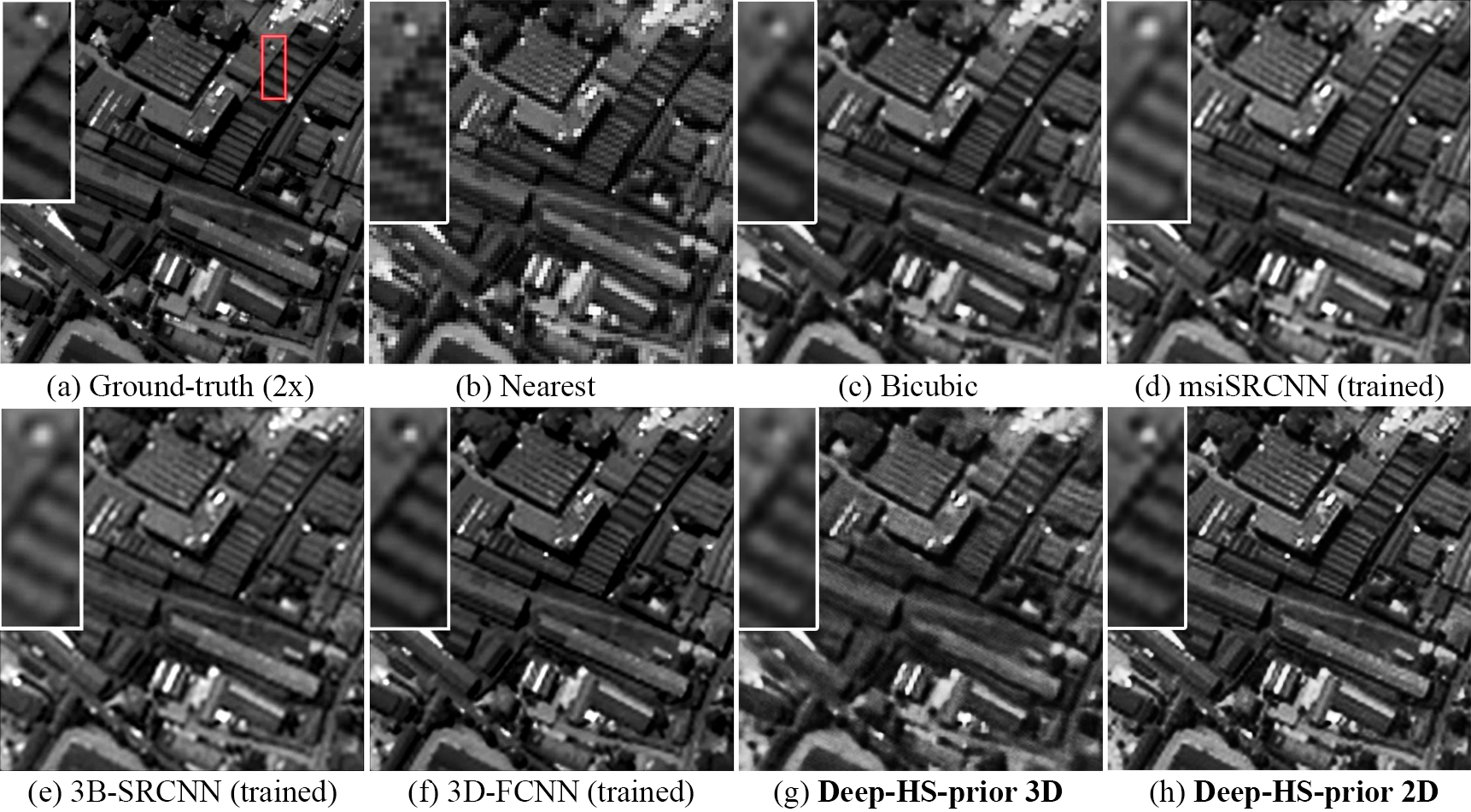}
\end{center}
   \caption{HSI super-resolution results. Pavia Center image; rescale factor 2; band 25 visualization.}
\label{fig:5}
\end{figure*}

\paragraph{Super-resolution.}
 The experiment was conducted using ROSIS-03 image of Pavia Center \cite{fauvel2012advances} (102 spectral bands). A patch of $150\times150$ pixels was cropped from the original image and downsampled by a factor of 2 by spatial dimensions. The evaluation includes ``nearest" and bilinear upsampling, learning-based method SRCNN \cite{dong2015image} applied band-wise (msiSRCNN) or by groups of 3 bands (3B-SRCNN \cite{liebel2016single}), and 3D-FCNN \cite{mei2017hyperspectral}. Results are presented in Fig. \ref{fig:5} and Table \ref{tab:1}.

\begin{table*}
\begin{center}
\begin{tabular}{|p{1.7cm}|p{1.7cm}p{1.7cm}p{1.7cm}p{1.7cm}p{1.7cm}p{1.7cm}p{1.7cm}|}
\hline
\multicolumn{8}{|c|}{Denoising} \\
\hline\hline
& HSSNR & LRTA & BM4D & LRMR & HSID-CNN &\textbf{Deep HS} & \textbf{Deep HS}\\
&       &      &      &      & (trained)&\textbf{prior 3D}&\textbf{prior 2D}\\
\hline
\textbf{MPSNR} & 16.31 & 23.17 & 22.57 & 24.31 & 25.29 & 23.24 & 25.05 \\
\textbf{MSSIM} & 0.605 & 0.849 & 0.812 & 0.879 & 0.901 & 0.852 & 0.889 \\
\textbf{SAM}   & 24.73 & 9.122 & 9.761 & 10.46 & 8.406 & 9.910 & 8.606 \\
\hline
\multicolumn{8}{c}{ }\\ 
\hline
\multicolumn{8}{|c|}{Inpainting} \\
\hline\hline
&  & Do    & Mumford- &TV-$H^{-1}$& FastHyIn&\textbf{Deep HS} & \textbf{Deep HS}\\
&  &Nothing& Shah   &           &         &\textbf{prior 3D}&\textbf{prior 2D}\\
\hline
\textbf{MPSNR} &  & 17.75 & 24.74 & 27.68 & 28.08 & 35.34 & 37.54 \\
\textbf{MSSIM} &  & 0.722 & 0.890 & 0.911 & 0.920 & 0.966 & 0.979 \\
\textbf{SAM}   &  & Inf & 5.429 & 3.855 & 3.032 & 1.133 & 0.856 \\
\hline
\multicolumn{8}{c}{ }\\ 
\hline
\multicolumn{8}{|c|}{Super-resolution} \\
\hline\hline
& Nearest&Bicubic&msiSRCNN&3B-SRCNN& 3D-FCNN &\textbf{Deep HS} & \textbf{Deep HS}\\
&    &    & (trained) & (trained) & (trained)&\textbf{prior 3D}&\textbf{prior 2D}\\
\hline
\textbf{MPSNR} & 29.98 & 31.10 & 32.48 & 32.69 & 33.92 & 32.31 & 33.67 \\
\textbf{MSSIM} & 0.921 & 0.937 & 0.957 & 0.960 & 0.969 & 0.945 & 0.967 \\
\textbf{SAM}   & 4.786 & 4.592 & 4.617 & 4.661 & 4.140 & 4.692 & 4.211 \\
\hline
\end{tabular}
\end{center}
\caption{Quantitative evaluation of the results.}
\label{tab:1}
\end{table*}

 \section{Results and discussion}
 As can be seen, the proposed method outperforms all single-image algorithms and demonstrate performance comparable to \emph{trained} CNNs, while \emph{not being trained} on any dataset before. Surprisingly, the 2D version of the Deep HS prior outperformed the 3D-convolutional one in all experiments. Note that the 2D implementation has nothing to do with band-wise processing; instead, it captures spectral information in filters of the first convolutional layer and uses combinations of them at the subsequent layers. Besides, the 3D implementation requires significantly more memory and computational time for processing the same amount of data because it works with tensors of higher dimensionality (roughly speaking a 3D version will have $C$ times more parameters, where $C$ is a number of bands). Eventually, we find 2D implementation was sufficient for processing of hyperspectral data.  And yet, it is worth mentioning that 3D version demonstrated comparable performance that is important from a theoretical point of view because it proves that CNN architectures based on 3D convolutions also contain the image prior within the intrinsic parameters.

\section{Conclusions}
Starting from the paradigm that the image prior can be found within a CNN itself and not be learned from training data or designed manually, we developed an effective single-hyperspectral-image restoration algorithm. Qualitative and quantitative evaluation of the results demonstrated superior effectiveness of the proposed algorithm compared to other single-image algorithms.

{\small
\bibliographystyle{ieee}
\bibliography{refs}
}

\end{document}